\documentclass[nobalancelastpage]{article}

\usepackage{microtype}
\usepackage{graphicx}
\usepackage{booktabs} 

\usepackage{tikz}

\usepackage{xcolor}
\usepackage{amsmath}
\usepackage{amsfonts}
\usepackage{caption}
\usepackage{float}
\usepackage{subcaption}
\usepackage{graphicx}
\usepackage{bbm}
\usepackage{bm}
\usepackage{listings}
\usepackage{stfloats}
\usepackage{pgfplots}
\usepgfplotslibrary{fillbetween}

\usetikzlibrary{bayesnet}

\renewcommand{\vec}[1]{\bm{\mathrm{#1}}}

\DeclareMathOperator{\hard}{hard}
\DeclareMathOperator{\sigmoid}{sigmoid}
\DeclareMathOperator{\KL}{KL}
\DeclareMathOperator{\ELBO}{ELBO}
\DeclareMathOperator{\Enc}{Enc}

\usepackage{hyperref}

 \usepackage[accepted]{icml2019}

\icmltitlerunning{SEALion}

\begin{document}

\twocolumn[
\icmltitle{SEALion: a Framework for Neural Network Inference on Encrypted Data}



\icmlsetsymbol{equal}{*}

\begin{icmlauthorlist}
\icmlauthor{Tim van Elsloo}{uva}
\icmlauthor{Giorgio Patrini}{deeptrace}
\icmlauthor{Hamish Ivey-Law}{data61}
\end{icmlauthorlist}

\icmlaffiliation{uva}{University of Amsterdam}
\icmlaffiliation{deeptrace}{Deeptrace; most work was done while at the University of Amsterdam}
\icmlaffiliation{data61}{CSIRO Data61}

\icmlcorrespondingauthor{Tim van Elsloo}{tim.van@elsl.ooo}

\icmlkeywords{deep learning, privacy, homomorphic encryption}

\vskip 0.3in
]



\printAffiliationsAndNotice{}  

\begin{abstract}

We present SEALion: an extensible framework for privacy-preserving machine learning with homomorphic encryption. %
It allows one to learn deep neural networks that can be seamlessly utilized for prediction on encrypted data.
The framework consists of two layers: the first is built upon TensorFlow and SEAL and exposes standard algebra and deep learning primitives; the second implements a Keras-like syntax for training and inference with neural networks. %
Given a required level of security, a user is abstracted from the details of the encoding and the encryption scheme, allowing quick prototyping.
We present two applications that exemplifying the extensibility of our proposal, which are also of independent interest: i) improving efficiency of neural network inference by an activity sparsifier and ii) transfer learning by querying a server-side Variational AutoEncoder that can handle encrypted data.



\end{abstract}


\section{Introduction}


%

Despite the success of machine learning prediction services, privacy of the client's data is seldom  considered a top priority in their deployment. However, when machine learning is applied to problems involving sensitive data, privacy is a requirement that cannot be eliminated.
An example is with medical applications \citep{kononenko2001machine}.
Assume that a hospital aims to automate diagnoses based on medical imagery. The hospital may not possess enough (labeled) data in order to train accurate predictive models. It could however collaborate with a third-party service providing such predictions, yet ethical and legal requirements need to be satisfied due to the sensitivity of patients' data.

Homomorphic Encryption (HE) is emerging as a framework for privacy preserving prediction services \citep{gilad2016cryptonets, bourse2018fast, sanyal2018tapas}. Upon  agreement on a protocol, the client --- the only private key holder --- encrypts and sends its data to the server. The server performs prediction in the encrypted domain and gains no knowledge of either input or any intermediate result. Finally, the client decrypts the predictions. With respect to privacy, the whole transaction can take place without the establishment of trust between the two parties --- as in a non-private prediction service. The client must still trust the server to generate correct predictions.


However, integrating homomorphic encryption with machine learning models is non-trivial and requires ad-hoc solutions. In addition, computation on encrypted data is orders of magnitude slower than the same operation on the original data. Given that it also requires careful selection of the encryption parameters, prototyping  models requires much more time and resources.


In this paper, we introduce SEALion, a new framework for implementing machine learning algorithms that can operate on homomorphically encrypted data and generate encrypted predictions. 
The design is largely inspired by the earlier work of CryptoNets \cite{gilad2016cryptonets}, and the SEAL library \cite{laine2016seal} based on the encryption scheme of \cite{fan2012somewhat}. Our framework aids in quickly prototyping multivariate polynomial functions on encrypted data in general, and their use in neural networks in particular, without requiring expert knowledge in the field of cryptography. A main ingredient of the framework is a heuristic search algorithm for selecting optimal parameters of the encryption scheme.

We showcase the use of SEALion with two applications of machine learning research. First, we demonstrate how to improve the speed of neural networks at inference time -- a current major drawback with homomorophic encryption. The most expensive operation of CryptoNets is the \emph{squared activation function}; we sparsify the network activations by the method of \cite{louizos2017learning}.

Second, we introduce the new scenario of \emph{encrypted transfer learning}. A client queries a server-side Variational Auto-Encoder, VAE \citep{kingma2013auto}, to obtain representations used for transfer learning in downstream tasks. The client is the private key holder: it sends an encrypted input to the server and obtains an encrypted representation, without sacrifycing the privacy of the input. 




\section{Background}


\subsection{Homomorphic Encryption}
\label{sec:homomorphic}

A homomorphism is a map $h: X \mapsto Y$ that preserves the algebraic structure of $X$ in its image in $Y$. That is, for any elements $a$ and $b$ in $X$, it is always true that $h(a + b) = h(a) \oplus h(b)$ and $h(a \cdot b) = h(a) \odot h(b)$, where $\oplus$ and $\odot$ represent addition and multiplication in $Y$. Consequently, for a multivariate polynomial $f$ with suitable coefficients, we have $h(f(x_1, \dots, x_n)) = f(h(x_1), \dots, h(x_n))$. In homomorphic encryption, $h$ is the encryption function, $X$ is the plaintext space, $Y$ the ciphertext space, and $f$ the function that we aim to evaluate on the encrypted data.

We use homomorphic encryption to conceal privacy-sensitive data without losing the ability to perform simple algebraic operations such as addition, subtraction and multiplication.
Our framework is based on SEAL~\cite{laine2016seal} and uses \textit{leveled} homomorphic encryption, 
which requires knowing the number of encrypted operations in advance. 
Related homomorphic schemes relax this limitation, but at the price of much higher computational cost associated with bootstrapping~\citep{chillotti2016faster}.


For details on the encryption scheme, we refer the reader to~\cite{fan2012somewhat,laine2016seal} and we will treat it as a blackbox with the following interface. Every message is encoded as an $n$-degree polynomial $\vec{m}$. Given a message polynomial $\vec{m}$, we have two functions $\textrm{Enc}(\vec{m})$ and $\textrm{Dec}(\vec{c})$ that satisfy the following equivalences for addition: $\vec{a} + \vec{b} = \textrm{Dec}\left(\textrm{Enc}(\vec{a}) \oplus \textrm{Enc}(\vec{b})\right)$ and for multiplication: $\vec{a} \cdot \vec{b} = \textrm{Dec}\left(\textrm{Enc}(\vec{a}) \odot \textrm{Enc}(\vec{b})\right)$. This naturally extends to subtraction and negation as well, but does not extend to division. Every operation on the message polynomial $\vec{m}$ is performed within the ring of integers modulo $t$ (the plaintext coefficient modulus), and every operation on the ciphertext polynomial $\vec{c}$ is performed within the ring of integers modulo $q$ (ciphertext coefficient modulus).

Every operation, including $\textrm{Enc}(\vec{m})$, introduces noise that is dependent on $\frac{t}{q}$ into the ciphertext polynomial $\vec{c}$. For example, after adding two ciphertexts together, the resulting ciphertext will have the sum of the noise of the original ciphertexts. The size of ciphertext polynomials is determined by $n$ and $q$. The ciphertext coefficient modulus $q$ is selected to provide a specific security level $\lambda$ (e.g.~128 bits). The SEAL library has a built-in $(n,\lambda) \mapsto q$ that maps a polynomial degree to a ciphertext coefficient modulus that is suitable for the requested security level. Finally, $t$ affects the level of noise that each operation introduces but also restricts the plaintext space.

With these primitives, we can evaluate multivariate polynomial functions on homomorphically encrypted data. The key to performance and accuracy is determining the most efficient $(n, q, t)$ for which we can still evaluate the function.

\subsection{Privacy-Preserving Inference with Homomorphic Encryption}
\label{sec:existing-work}

Homomorphic primitives are applied in machine learning applications for privacy preservation. Our focus is on our particular client-server scenario: the server trains a model on non-sensitive data; the client (the private key holder) queries the server to perform inference, e.g. to classify an image, after encrypting its sensitive data; the result is computed homomorphically as a ciphertext by the server; only the client can decrypt it, preserving privacy of both input and output of the service.

A different scenario is that of Secure Multi-party Computation (SMC), where the involved parties evaluate a model jointly. While SMC can compute arbitrary functions, it needs to involve multiple party for privacy preservation and its security model is different from ours. In the following, we mainly cite previous work that is relevant to our setting.

\cite{gilad2016cryptonets} introduce CryptoNets, a class of convolutional neural networks optimized for high throughput, that use square activations and input discretized within an arbitrary interval. The leveled homomorphic encryption scheme of \cite{bos2013improved,laine2016seal} is used, which implies that the depth of the neural network is fixed beforehand, depending on the encryption parameters and architectural choices. The encryption $(n, q, t)$ are manually chosen for good performance. Our contribution is built upon CryptoNets by providing a modular and extensible software architecture\footnote{The SEALion library will be open-sourced at publication time.}, and automatic parameters selection. These features are demonstrated by the straightforward implementation of neural networks architectural sparsity and of Variational AutoEncoders in this work.
Faster Cryptonets are introduced by~\citep{chou2018faster}, which uses pruning and quantization to reduce the computational complexity of the network.

\cite{bourse2018fast} propose a class of neural networks optimized
for low latency, using sign activation functions and binarized input. It uses a fully homomorphic encryption inspired by \citep{chillotti2016faster} and can theoretically construct networks of arbitrary depth. In fact, the authors observe that the sign
function coincidentally can be performed as part of a  bootstrapping procedure, which may rid of the noise in the encrypted numbers, allowing deeper networks. However, their use of boostrapping implies stochasticity and the sign activations introduce additional training issues, limiting the practice to fairly simple neural networks.

In contrast, \cite{sanyal2018tapas} develop fully homomorphic encryption to evaluate a network by performing every arithmetic operation as a composition of binary gates. Every operation consists of many bootstraps and is therefore inherently immune to noise and has no architectural restrictions. However, it is slower than any previous work: one prediction on MNIST takes 32 hours on a single core.

Outside our security model, a hybrid of homomorphic encryption and multi-party computation is used for training and inference in \cite{hardy2017private}. \cite{juvekar2018gazelle} propose a pragmatic hybrid scheme of homomorphic encryption and garbled circuits, in which the client actively assists in the privacy-preserving evaluation of a neural network. 
In the approach by~\cite{makriepic}, the model owner's SVM classifier is trained on features extracted with a pre-trained deep convolutional network; subsequently, data owners can encrypt and submit their features to the SVM.
In~\cite{juvekar2018gazelle}, a model was evaluated using garbled circuits, which requires active participation of the client, beyond merely encrypting and decrypting the data.

Recently, two frameworks for privacy-preserving deep learning with secure multi party computation have been introduced by \cite{ryffel2018generic, dahl2018private}, respectively building on top of PyTorch an TensorFlow. They share a similar intent of ours, focussing on usability and extensibity, but implement SMC and do not use homomorphic encryption.

A low-level compiler framework for homomorphic encryption is detailed in~\citep{dathathri2018chet}. By traversing programs that run on encrypted data during the compilation procedure, it can transform programs into optimized homomorphic evaluation circuits.

\subsection{Discretized neural networks}

Due to the fact that all operations on ciphertexts are performed in the ring of integers modulo $q$, we are restricted to integer weights, rather than usual floating point.
In order to obtain gradients with respect to 
discretized weights, one can use the straight-through
estimator \citep{bengio2013estimating, courbariaux2016binarized}: gradients for backpropagation are simply approximated by the original ones, where no discretization is in place.
In contrast \cite{bourse2018fast} round weights after training, introducing errors;
it is unclear how this is achieved by \cite{gilad2016cryptonets}.




%






\section{Design}

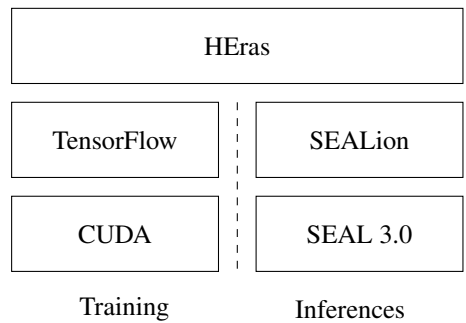
\begin{figure}[t]
    \centering
    \begin{tikzpicture}
        \draw[black] (0, 0) rectangle (6, 1);
        \draw node at (3, 0.5) {HEras};
        \draw[black] (0, -1.25) rectangle (2.75, -0.25);
        \draw node at (1.375, -0.75) {TensorFlow};
        \draw[black] (3.25, -1.25) rectangle (6, -0.25);
        \draw node at (4.625, -0.75) {SEALion};
        \draw[black] (0, -2.5) rectangle (2.75, -1.5);
        \draw node at (1.375, -2) {CUDA};
        \draw[black] (3.25, -2.5) rectangle (6, -1.5);
        \draw node at (4.625, -2) {SEAL 3.0};
        
        \draw[dashed] (3, -0.25) -- (3, -2.5);
        
        \draw node at (1.5, -3) {Training};
        \draw node at (4.5, -3) {Inferences};
    \end{tikzpicture}

    \caption{Architecture of our framework.}
    \label{fig:overview}
\end{figure}

An overview of the architecture of our framework is shown in fig. \ref{fig:overview}. SEALion itself only implements primitive objects and operations (sec. \ref{sec:primitives}), is not focused on machine learning and therefore has no notion of training or inference. Instead, it can only be used to evaluate multivariate polynomial functions on batches of encrypted data (sec. \ref{sec:batching}), optionally using CRT to expand the input and output domain of those functions (sec. \ref{sec:crt}). On top of SEALion, we have built another layer called HEras, that provides an abstraction over training neural networks on plaintext data using TensorFlow, and running inference on encrypted data using SEALion.

SEALion's internal engine is inspired by TensorFlow. Polynomial functions are expressed using a computational graph. Each node in this graph represents either a \textit{placeholder} for some input data, or the result of some \textit{operation}, and has an associated estimated domain. With those domains, this graph aids in automatically selecting the most efficient encryption that can fit the entire in- and output domain of the polynomial function.

\subsection{SEALion: Primitives}
\label{sec:primitives}

SEALion exposes many primitives that can be applied to plaintext and ciphertexts. It supports addition, subtraction, negation and multiplication. Each of those operations take either a single ciphertext and a plaintext, or two ciphertexts. In addition, our framework also support element-wise addition and multiplication of tensors, and matrix multiplication and inner products.

\subsubsection{Batching}
\label{sec:batching}

Like~\cite{gilad2016cryptonets}, SEALion also supports batching to minimize the amortized runtime. By default, encoded in a single plaintext polynomial is a single integer that can be approximately $n \log_2 t$ bits. Instead, we can also use SIMD~\citep{smart2014fully} to encode $n$ smaller numbers of $\log_2 t$ bits, $-\frac{t}{2} \leq a \leq \frac{t}{2}$ in a single polynomial. For this to work, $t$ needs to be prime and $t \equiv 1 \pmod{2n}$.

SEALion automatically takes care of selecting encryption parameters that support batching and the packing of ciphertexts. In particular, batches provided to SEALion are automatically zero-padded into the batch size supported by the encryption parameters.

\subsubsection{Chinese Remainder Theorem}
\label{sec:crt}

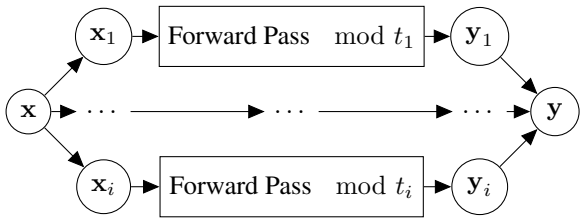
\begin{figure}[t]
    \centering
    \begin{tikzpicture}
        \node[circle, draw] at (-1, -1) (x) {$\vec{x}$};
        \node[circle, draw] at (0, 0) (x1) {$\vec{x}_1$};
        \node[circle, draw] at (0, -2) (xi) {$\vec{x}_i$};
        
        \node[rectangle, draw, minimum height=0.8cm, minimum width=3.5cm] at (2.5, 0) (ff1) {Forward Pass $\mod t_1$};
        \node[rectangle, draw, minimum height=0.8cm, minimum width=3.5cm] at (2.5, -2) (ffi) {Forward Pass $\mod t_i$};
        
        \node at (0, -1) (xd) {$\dots$};
        \node at (2.5, -1) (ffd) {$\dots$};
        \node at (5, -1) (yd) {$\dots$};
        
        \node[circle, draw] at (5, 0) (y1) {$\vec{y}_1$};
        \node[circle, draw] at (5, -2) (yi) {$\vec{y}_i$};
        
        \node[circle, draw] at (6, -1) (y) {$\vec{y}$};
        
        \draw (x) edge[->] (x1);
        \draw (x) edge[->] (xd);
        \draw (x) edge[->] (xi);
        
        \draw (x1) edge[->] (ff1);
        \draw (xd) edge[->] (ffd);
        \draw (xi) edge[->] (ffi);
        
        \draw (ff1) edge[->] (y1);
        \draw (ffd) edge[->] (yd);
        \draw (ffi) edge[->] (yi);
        
        \draw (y1) edge[->] (y);
        \draw (yd) edge[->] (y);
        \draw (yi) edge[->] (y);
    \end{tikzpicture}
    \caption{Overview of using CRT to decompose large integers during the forward pass into smaller components that can be evaluated in parallel.}
    \label{fig:crt}
\end{figure}

Recall that larger plaintext coefficient moduli adversely affect the remaining noise budget after each operation. In order to evaluate functions of increasing depth, we instead resort to the Chinese Remainder Theorem to decompose our large plaintext coefficient modulus into smaller moduli. Specifically, we evaluate the function once for each smaller modulus, and conclude by assembling the outputs into a single number using the CRT (fig. \ref{fig:crt}). This approach was also taken by~\cite{gilad2016cryptonets}.

\subsection{HEras: Neural Networks}

\label{sec:discretization}

\begin{figure*}[t]
\small
    \begin{lstlisting}
model = Sequential()
model.add(Flatten(input_shape = (28, 28, 1), input_saturation = 4))
model.add(Dense(units = 100, saturation = 2 ** 4))
model.add(Activation())
model.add(Dense(units = 10, saturation = 2 ** 4))

model.compile(loss = 'categorical_crossentropy', optimizer = 'adam',
              metrics = [ 'categorical_accuracy' ])

# Train the neural network on plaintext data.
model.fit(x_train, y_train, epochs = 10, batch_size = 64)
    \end{lstlisting}
    \caption{An image classifier built with HEras.}
    \label{fig:example-training}
\end{figure*}

\begin{figure*}[t]
\small
    \begin{lstlisting}
# 1. Encrypt the test set. Parameters have been selected during training.
pk, sk   = sl.Keypair(model.encryption_params)
e_x_test = pk.encrypt(x_test)

# 2. Run the model on the encrypted input.
e_p_test, metrics = model.predict(e_x_test, encrypted = True)

# 3. Decrypt the predictions and use argmax to get predicted classes.
p_test = sk.decrypt(e_p_test)
p_test = np.argmax(p_test, axis = -1)
    \end{lstlisting}
    \caption{Encrypted inference with the image classifier shown in fig. \ref{fig:example-training}.}
    \label{fig:example-inference}
\end{figure*}

A second layer, HEras (Homomorphic Encryption Keras), is built on top of these primitives: a neural network framework with an interface similar to Keras, but that can operate on homomorphically encrypted data. During training, the parameters of a model are learned with TensorFlow. Once trained, the model parameters are copied to the inference engine that runs on top of SEAL.

On a lower level, every \textit{forward} pass through the network is implemented in two-fold: one implementation is used during training and built on top of TensorFlow and is used for learning the weights of the network, the other implementation is used during inference for generating predictions on encrypted data calling SEALion primitives. Every operation in the \textit{forward} pass (except for the final activation function, and the loss function) must be a polynomial function with integer coefficients. The \textit{backward} pass is used in backpropagation for learning the weights -- training never required handling encrypted data.

All built-in layers take care of discretization, by using straight-through estimators to obtain gradients for the discretization function $w' = \lfloor w \cdot s \rceil$ where $w$ are the original floating point weights, and $s$ is a user-provided saturation hyper-parameter. Higher $s$ will lead to increasingly saturated weights but will likely also increase the magnitude of the model outputs and thus require more plaintext coefficients (sec. \ref{sec:crt}).

Under the hood, the \textit{forward} pass of a neural network in HEras yields a single multivariate polynomial, that is evaluated like any other polynomial in SEALion.
In fig. \ref{fig:example-training}, we show an example of a training procedure built with HEras. In particular, observe the required saturation levels that restrict the input domain and aid in estimating the output domain of the underlying prediction function. In fig. \ref{fig:example-inference}, we show an example of inference with encrypted data.

HEras runs Adam~\citep{kingma2014adam} by default due to a its per-layer adaptive learning rate, which aids in overcoming the training issues linked to the square activations for deep networks.

\subsection{Automatic Encryption Parameters Selection}


The plaintext modulus $t$ is estimated by traversing the graph and computing the output domain based on the domains of each intermediate operation. When using HEras, the estimate can also be computed by evaluating the computational graph on a small batch of plaintext training data, and recording the output domain. Since the test set might follow a different distribution, we add a variable margin to the recorded domain. The plaintext modulus is subsequently estimated to be $t = 2 \cdot \max\left(|l|, |u|\right) \cdot \lambda$ where $l$ and $u$ are the lower- and upperbound of the domain respectively and $\lambda \geq 1$ is some margin.

Then, we use the CRT to split the modulus in smaller moduli that are easier to compute. For this to work, the smaller moduli need to be coprime numbers. In addition, for batching to work, each moduli must satisfy $(t_i \bmod 2n) = 1$, where $t_i$ is one of the moduli and $n$ is the polynomial degree.

Next, we compare pairs $(n, t)$ of polynomial degrees and plaintext moduli. The ciphertext coefficient modulus is provided by SEAL and does not depend on the selected plaintext coefficient modulus or vice versa. Empirically, we have found that the runtime of each operation depends linearly on the polynomial degree $n$. However, we can pack more data in polynomials of a larger degree and thus achieve lower amortized runtime. In addition, runtime depends linearly on the number of plaintext moduli $t$ (see sec. \ref{sec:crt}). In sum, our heuristic can either optimize for latency by selecting a candidate with the lowest polynomial degree $n$, or for amortized throughput by selecting a candidate with a higher polynomial degree but fewer plaintext moduli $t$. The list of candidates is sorted by this objective.

Finally, the multivariate polynomial function is evaluated on a small training set for each parameters candidate and we select the first candidate that can successfully evaluate the function without running out of noise budget. This occurs when one of the plaintext moduli $t_i$ is too large, since the noise growth is strongly dependent on $t_i$. In that case, we decompose $t$ in more but smaller $t_i$ and try again.

The security level is a parameter that must be provided by the user; the automatic selection of encryption parameters guarantees that the security level is achieved.


\subsection{Implementation}

SEAL's APIs are exposed to Python via a C++ bridge. On top of that bridge, the primitives layer takes care of abstracting away the notion of dealing with multiple smaller plaintext moduli and handling tensors in addition to scalars.
The neural network framework is written entirely in Python and resembles the Keras API. However, it does not support non-polynomial functions (such as ReLU or sigmoid) that are unavailable when using homomorphic encryption.

\section{Application: Sparsification of Neural Network Architectures}

In our first case study, we show that our framework can be extended to handle model sparsification, with the technique of~\cite{louizos2017learning}. An $L_0$ regularizer is applied to sparsify the squared activations, the main performance bottleneck of SEALion's deep neural networks. Sparsifying the weights instead of the activations would not achieve maximal speedup, because computing the square of ciphertexts is much more expensive than multiplying plaintexts with ciphertexts.

\subsection{Background}

The number of non-zero parameters in a neural network can be seen as an $L_0$ norm. Using this norm directly to enforce sparsity is non trivial, due to its intrinsic non-differentiability.
\cite{louizos2017learning} present a solution based on a smooth approximation. Each network parameter $\theta$ is multiplied with a binary gate, i.e. $
\theta \cdot z$. 
In order to obtain a differentiable parametrization of the network, we can sample from a binary concrete
distribution \citep{maddison2016concrete} with parameters $(\log \alpha, \beta)$, stretch on the interval $(\zeta, \gamma)$ and finally pass through a hard-sigmoid:
$\textrm{hard}(\cdot) = \max(0, \min(1,
\sigmoid(\cdot) \left(\zeta - \gamma\right) + \gamma))$. In sum, during training each gate $z$ is sampled according to eq. \ref{eq:l0-gate}
where $u \sim \mathcal{U}(0, 1)$:
\begin{align}
z = \hard\left(\log\left(\frac{u}{1 - u} + \alpha \right) / \beta\right)
\label{eq:l0-gate}
\end{align}
The parameters $(\log \alpha, \beta)$ are learned with the reparametrization trick \citep{kingma2013auto}. An approximation of the $L_0$ norm is then obtained as $\sum_{j=1}^{|\theta|}
\sigmoid(\log \alpha_j - \beta \log \frac{-\gamma}{\zeta})$ and used as regularizer. At inference,
deterministic gates are computed as:  
$\hard\left(\log \alpha\right)$ and rounded.
The same technique can be applied for sparsifying activations by sharing a gate with multiple parameters.

\subsection{Implementation}

During training, after computing the output of a convolutional layer, the feature maps are multiplied with $\left\lfloor z \right\rceil$, where we use the same straight-through estimator to obtain a gradient for the discretization function as we did for other parameters (see sec. \ref{sec:discretization}). We add $\sum \textrm{sigmoid}\left(\alpha - \beta \log \frac{-\gamma}{\zeta}\right)$ as a regularization to the loss function.

For inference, we compute $\hat{z} = \hard\left(\log \alpha\right)$ and pass $\left\lfloor \hat{z} \right\rceil$ to SEALion when performing the convolution on the encrypted ciphertexts. In turn, SEALion will skip computing values in the resulting feature maps for which $\left\lfloor \hat{z} \right\rceil = 0$, and therefore skip the computation of several activations.
Users of HEras can pass the built-in $L_0$ activity regularizer as an argument to the convolutional layer constructor.

\subsection{Experiments}

\begin{table*}[t!]
    \centering
    \small
    \begin{tabular}{lcccccc}
        \toprule
        \textbf{Model} & \textbf{Parameters} & \textbf{Activations} & $n, |t|$ & \textbf{Lat.}[s] & \textbf{TP}[1/h] & \textbf{Acc. }[\%] \\
        \midrule
        \textit{Earlier Work} \\
        CryptoNets & 126,375 & 945 & $4096, 2$ & 250 & 58,982 & 98.95 \\
        Faster CryptoNets & --- & 945 & $8192, 2$ & 39.1 & 754,250 & 98.71 \\
        DiNN-30 & 26,520 & 30 & --- & 0.491 & 6,990 & 93.46 \\
        DiNN-100 & 79,400 & 100 & --- & 1.64 & 2,143 & 96.35 \\
        TAPAS & --- & --- & --- & 32[h] & --- & 98.60 \\
        \midrule
        \textit{Our Work} \\
        DNN-30 & 26,520 & 30 & $4096, 1$ & 1.14 & 12,933,344 & 97.40 \\
        DNN-100 & 79,400 & 100 & $4096, 1$ & 3.28 & 4,494,965 & 98.01 \\
        CNN-$16$ & 79,800 & 3236 & $8192, 2$ & 192 & 153,600 & 98.96 \\
        CNN-$16$-$L_0$ & 79,800 & 762 & $8192, 2$ & 60 & 491,520 & 98.91 \\
        \bottomrule
    \end{tabular}
    \caption{Our baselines (DNN-30, DNN-100) and our convolutional networks with and without $L_0$ activity sparsifier in comparison to earlier work by~\cite{gilad2016cryptonets,chou2018faster,bourse2018fast,sanyal2018tapas}. The number of plaintext moduli is denoted with $|t|$ (see sec. \ref{sec:crt}) and is only reported for models built on~\cite{laine2016seal}. Other metrics in order: latency, throughput and accuracy. Runtime is not strictly comparable with previous work due to differences in the hardware and, in same cases, in the security parameters (see text). For comparison, a standard neural network with ReLU activations and no discretization, with an architecture as our CNN-$16$ achieves $98.84\%$ and can generate 120M predictions per hours on plaintext input.}
    \label{fig:overall}
\end{table*}

Our model consists of a 1) convolutional layer with 16 filters of size $5 \times 5$ and $2 \times 2$ strides, 2) a square activation layer, 3) a pooling layer with kernel size $3 \times 3$ and again $2 \times 2$ strides, 4) a fully connected layer with 100 units and finally 5) the output layer with 10 units. The $L_0$ activity regularizer is applied to the filter maps of the convolutional layer.

This architecture is different from~\citep{gilad2016cryptonets,chou2018faster}: we use more convolutional filters but omit their second convolutional and pooling layers. In total, the number of non-linear layers is the same.

We evaluate the performance of our models on MNIST~\citep{lecun1998gradient}, because it was also used in earlier work that we can compare with~\cite{gilad2016cryptonets,chou2018faster,bourse2018fast,sanyal2018tapas}.

In table \ref{fig:overall}, we compare with the prior state of art. We have implemented two non-convolutional baseline models with the same architecture as DiNN-$30,100$ from~\citep{bourse2018fast}. Our MLP-$30, 100$ models have approximately $2000$ times higher throughput and higher accuracy than DiNN-$30,100$ respectively, but twice the latency.

\begin{figure}[b]
    \centering
        \begin{tabular}{cccccc}
            \textbf{Epoch} & \textbf{1} & \textbf{4} & \textbf{16} & \textbf{64} & \textbf{128} \\
        \midrule
            &
            \includegraphics[width=.1\linewidth]{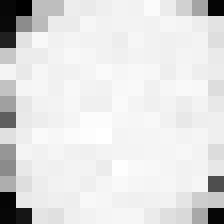} &
            \includegraphics[width=.1\linewidth]{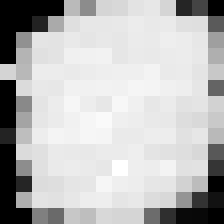} &
            \includegraphics[width=.1\linewidth]{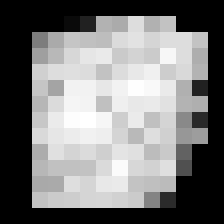} &
            \includegraphics[width=.1\linewidth]{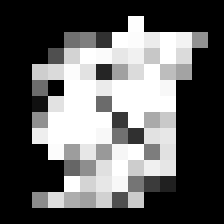} &
            \includegraphics[width=.1\linewidth]{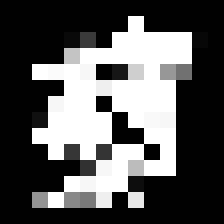} \\
            &
            \includegraphics[width=.1\linewidth]{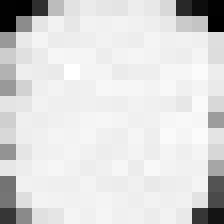} &
            \includegraphics[width=.1\linewidth]{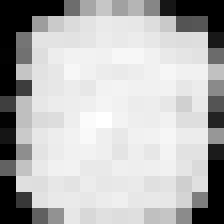} &
            \includegraphics[width=.1\linewidth]{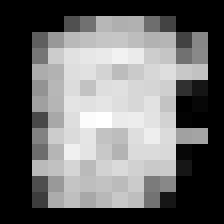} &
            \includegraphics[width=.1\linewidth]{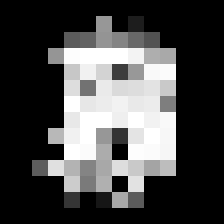} &
            \includegraphics[width=.1\linewidth]{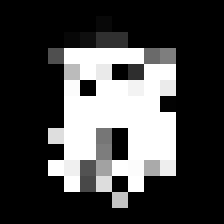} \\
    \end{tabular}
    \caption{Evolution of gates on two feature maps extracted from CNN-$16 + L_0$ trained on MNIST.}
    \label{fig:l0-evo}
\end{figure}

Our CNN-$16$ is considerably slower than our non-convolutional models, due the higher number of activations, but it obtains the same accuracy as CryptoNets~\cite{gilad2016cryptonets}. It improves almost 25\% on latency and achieves almost 3 times higher throughput. This is due to the architectural differences.

Our CNN-$16$-$L_0$ model is based on CNN-$16$ but sparsifies the model from 3236 activations down to just 762, while preserving the same level of accuracy. The sparsified model is more than twice as fast as the original model, and has almost 10 times higher throughput than CryptoNets.

In fig. \ref{fig:l0-evo} we show the evolution of the gates that mask the feature maps of CNN-$16$-$L_0$ over time during training. The regularizer discards information deemed irrelevant for prediction, such as the MNIST black borders.

All of our models satisfy 192 bits of security and we borrow $\texttt{coeff\_modulus\_192}(n)$ from SEAL to select the ciphertext coefficient moduli $q$, where $n$ is the polynomial degree. For our DNN-30 and DNN-100 models, $(n, \log_2 q) = (4096, 75)$ and for our CNN-$16$ and CNN-$16$-$L_0$ models, $(n, \log_2 q) = (8192, 152)$. We have verified the security level of these parameters per the recommendations established by Homomorphic Encryption Standardization Workshop~\citep{chase2017security}. We are comparing with CryptoNets, $(n, \log_2 q) = (4096, 191)$, for which we do not have an accurate security estimate, but in general lower $q$ for the same polynomial degree $n$ raises the security level~\citep{dowlin2017manual}.




\section{Application: Encrypted Transfer Learning with Discretized VAE}

In our second application of SEALion, we describe the novel setting of \emph{encrypted transfer learning}.
Specifically, owners of small datasets can exploit existing knowledge from cloud services that expose embedding obtained by unsupervised learning on similar (but unlabeled) datasets ~\citep{bengio2012deep,noroozi2016unsupervised,pathak2016context}.
The expected outcome is a richer representation of the client data that can be transferred to allow the client to learn models downstream, or to perform other data analysis tasks, without sacrificing privacy. As for classification, only the client has access to plaintext input and output of the service.

We apply Variational AutoEncoders \citep{kingma2013auto,rezende2014stochastic} for unsupervised learning by the server.

\begin{figure*}[t!]
\tiny
    \input{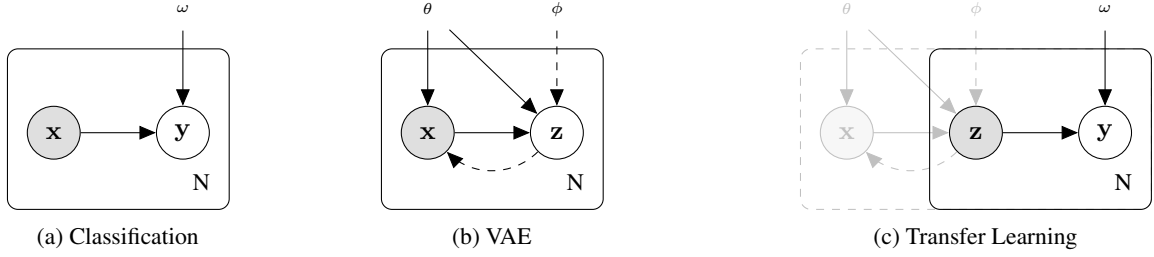}
    \caption{Graphical models of a classifier, the Variational AutoEncoder and its application in our
             transfer learning protocol.}
    \label{fig:diff-vae}
\end{figure*}

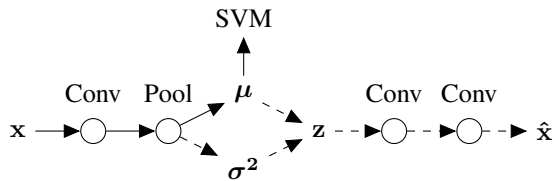
\begin{figure}[b]
\centering
\begin{tikzpicture}
\node (x) {$\vec{x}$};

\node[draw, circle] at (1, 0) (conv) {};
\node at (1, 0.5) {Conv};

\node[draw, circle] at (2, 0) (pool) {};
\node at (2, 0.5) {Pool};

\node at (3, 0.5) (mu) {$\vec{\mu}$};
\node at (3, -0.5) (sigma) {$\vec{\sigma^2}$};

\node at (3, 1.5) (svm) {SVM};

\node at (4, 0) (z) {$\vec{z}$};

\node[draw, circle] at (5, 0) (conv2) {};
\node at (5, 0.5) {Conv};

\node[draw, circle] at (6, 0) (conv3) {};
\node at (6, 0.5) {Conv};

\node at (7, 0) (xp) {$\vec{\hat{x}}$};

\draw (x) edge[->] (conv);
\draw (conv) edge[->] (pool);
\draw (pool) edge[->] (mu);
\draw (pool) edge[dashed, ->] (sigma);
\draw (mu) edge[dashed, ->] (z);
\draw (mu) edge[->] (svm);
\draw (sigma) edge[dashed, ->] (z);
\draw (z) edge[dashed, ->] (conv2);
\draw (conv2) edge[dashed, ->] (conv3);
\draw (conv3) edge[dashed, ->] (xp);

\end{tikzpicture}
\caption{Overview of our VAE implementation. Dashed lines represent layers with floating point parameters. Opaque lines represent layers that are evaluated on encrypted data and use straight-through estimators to obtain gradients for discretized parameters.}
\label{fig:vae-impl}
\end{figure}

\subsection{Background}

VAE is a latent variable model of the distribution of $\vec{x}$, i.e. $p(\vec{x}) = p(\vec{z}) p_\theta(\vec{x} | \vec{z})$.
VAE learns the parameters $\theta$ of the
generative model, assuming $p(\vec{z})$ as a Normal prior. However, since $\vec{z}$ is unobserved the true posterior
$p(\vec{z} | \vec{x})$ is intractable.

VAE takes a variational approximation of the true posterior as a parametric Gaussian $q_\phi(\vec{z} |
\vec{x}) = \mathcal{N}\left(\vec{z}; \vec{\mu},
{\vec{\sigma}^{2}} I\right)$ and maximize the ELBO, or Evidence Lower BOund of the marginal log-likelihood:
\begin{align}
& \mathbb{E}_{\vec{x}} \log p_{\theta}(\vec{x}) \geq  \ELBO = \nonumber \\
& \mathbb{E}_{\vec{x}} \left[ \mathbb{E}_{\vec{z} \sim q_\phi(\vec{z} | \vec{x})} \log p_{\theta}(\vec{x} | \vec{z}) - \KL(q_\phi(\vec{z} | \vec{x}) || ~p(\vec{z})) \right]. \nonumber 
\end{align}

During training, VAE jointly optimizes $\theta$ and $\phi$ by learning the generative model $p_\theta(\vec{x} | \vec{z})$, the decoder, and the variational approximation of the posterior $q_\phi(\vec{z} | \vec{x})$, the encoder, that outputs mean $\vec{\mu}$ and variance $\vec{\sigma}^2$ vectors (Figure \ref{fig:diff-vae-server}). Crucially, VAE introduces the reparametrization trick to allow efficient gradient estimation for $\phi$.
Sampling from $q_\phi(\vec{z} | \vec{x})$, i.e. embedding data with the encoder, yields a new representation useful for tasks such as semi-supervised and transfer learning \citep{kingma2014semi}.


\subsection{Implementation}

VAE consists of encoder and decoder models. During training, the encoder samples $\vec{z} \sim \mathcal{N}(\vec{\mu}, \vec{\sigma}^2)$ and the decoder attempts to reconstruct $\vec{x}$ from the latent representation $\vec{z}$, subject to the $\textsc{KL}$-term as regularizer. During inference, we do not need the decoder at all, and only use the latent representation for transfer learning --- in particular, we do not sample from the approximated posterior, but only compute the Normal mean $\vec{z} = \vec{\mu}(\vec{x})$. 

In fig. \ref{fig:vae-impl}, we show an overview of the VAE architecture.
Only the convolutional layer of the encoder model and the fully connected layer towards $\vec{\mu}$ are discretized. As such, only the path from $\vec{x}$ to $\vec{\mu}$ is evaluated during inference (the variance function is not used). The decoder does not require adaptations to HE and thus we can learn its weights $\theta$ in full precision, and use standard ReLU activations and batch normalization. We call our model dVAE, for ``discretized'', where discretization is applied to the encoder weights $\phi$.

In our protocol, the server first trains a VAE on a large
unlabeled dataset (fig. \ref{fig:diff-vae-server}). Instead of training a classifier directly on the small local dataset (fig. \ref{fig:diff-vae-supervised}), the client now sends encrypted data $\Enc(\vec{x})$. The server responds by performing encrypted inference 
$\Enc(\vec{z}) =  \vec{\mu}(\Enc(\vec{x}))$. Finally, the client decrypts $\Enc(\vec{z})$ with the private key to obtain the embeddings. Instead of training a local model of the original raw input $\vec{x}$, the client can now exploit the richer representation $\vec{z}$ and train a local classifier (fig. \ref{fig:diff-vae-client}). This is in contrast to the more usual framework (assumed in the rest of the paper) where the server exposes a classifier.

\subsection{Experiments}

\begin{figure}[t]
    \centering
    \small
        \begin{tabular}{c|cc|cc}
        \toprule
    \multicolumn{1}{c}{} & \multicolumn{2}{c}{\textbf{overlap}} & \multicolumn{2}{c}{\textbf{no overlap}} \\
        \multicolumn{1}{c}{\textbf{Originals}} &
        \multicolumn{1}{c}{\textbf{VAE}} &
        \multicolumn{1}{c}{\textbf{dVAE}} &
        \multicolumn{1}{c}{\textbf{VAE}} &
        \multicolumn{1}{c}{\textbf{dVAE}}\\
        \midrule
        \includegraphics[interpolate=off]{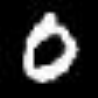} &
        \includegraphics[interpolate=off]{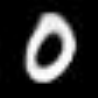} &
        \includegraphics[interpolate=off]{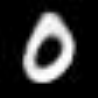} &
        \includegraphics[interpolate=off]{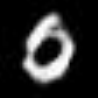} &
        \includegraphics[interpolate=off]{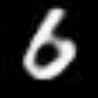} \\
        \includegraphics[interpolate=off]{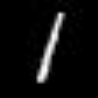} &
        \includegraphics[interpolate=off]{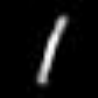} &
        \includegraphics[interpolate=off]{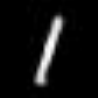} &
        \includegraphics[interpolate=off]{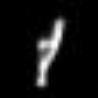} &
        \includegraphics[interpolate=off]{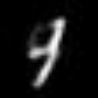} \\
        \includegraphics[interpolate=off]{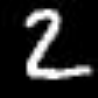} &
        \includegraphics[interpolate=off]{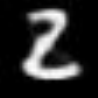} &
        \includegraphics[interpolate=off]{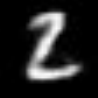} &
        \includegraphics[interpolate=off]{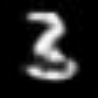} &
        \includegraphics[interpolate=off]{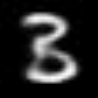} \\
        \bottomrule
    \end{tabular}
    \caption{Left: Reconstructions of MNIST digits by VAE and dVAE. Two posterior samples per image are given. Right: same but with models trained on classes 3-9 only and reconstructions from classes 0-2, simulating transfer learning.}
    \label{fig:vae-recon}
\end{figure}

We test dVAE on MNIST~\citep{lecun1998gradient} and EMNIST~\citep{cohen2017emnist}.
We test two variants of transfer learning. In both, the client trains a local Gaussian kernel SVM classifier on a small set of labelled data. In the overlapping setting, the SVM is trained on the same class distribution as the dVAE; e.g. for MNIST, they are both trained on classes 0-9. In the non-overlapping setting, the dVAE is trained on a subset of the classes and used for transfer on the remaining; e.g. for MNIST, it is trained on classes 3-9 and the classifier is learned on the representation of 0-2.

As an initial qualitative assessment, fig. \ref{fig:vae-recon} shows the reconstructions of dVAE and VAE.
The quality of the reconstructions of dVAE does not look too dissimilar than of regular VAE and therefore we assume that the underlying latent space is also not significantly degraded.


\begin{figure}[t!]
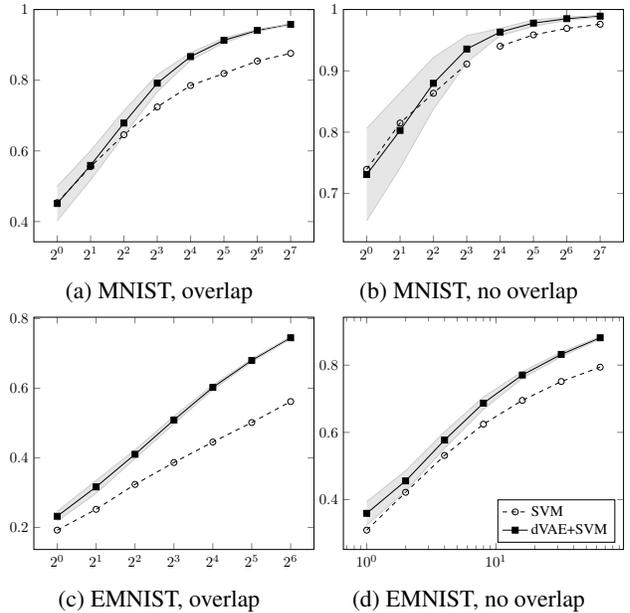

\centering
\begin{subfigure}[t]{.24\textwidth}
\resizebox{\textwidth}{!}{\begin{tikzpicture}
	\begin{axis}[
    xmode=log, legend pos=south east,
    ymax=1,
    width=2\textwidth,
    log basis x={2},
        legend style={font=\fontsize{3}{5}\selectfont},
	scatter/classes={%
		a={mark=square*,blue},%
		b={mark=triangle*,red},%
		c={mark=o,draw=black}}]
	
	\addplot[black,dashed,mark=o,mark options={solid}] coordinates {
	    (1, 0.453656)
	    (2, 0.555432)
	    (4, 0.64566)
	    (8, 0.7243200000000001)
	    (16, 0.7849720000000001)
	    (32, 0.8190559999999999)
	    (64, 0.8541640000000001)
	    (128, 0.876072)
	};
	\addlegendentry{SVM}
	\addlegendentry{+dVAE}
	\legend{}; 
    \input{sections/results/vae/cvae-mnist-overlap.tex}
	\end{axis}
\end{tikzpicture}}
\caption{MNIST, overlap} 
\end{subfigure}%
\begin{subfigure}[t]{.24\textwidth}
\resizebox{\textwidth}{!}{\begin{tikzpicture}
	\begin{axis}[
    xmode=log, legend pos=south east,
    ymax=1,
    width=2\textwidth,
    log basis x={2},
	scatter/classes={%
		a={mark=square*,blue},%
		b={mark=triangle*,red},%
		c={mark=o,draw=black}}]
	
	\addplot[black,dashed,mark=o,mark options={solid}] coordinates {
	    (1, 0.7394343819510644)
	    (2, 0.8148840165236733)
	    (4, 0.8634509056244042)
	    (8, 0.9111280584683825)

	    (16, 0.940197013028281)
	    (32, 0.9586018430251032)
	    (64, 0.9691897044804576)
	    (128, 0.9760279631394979)
	};
	\addlegendentry{SVM}
	\addlegendentry{dVAE+SVM}
	\legend{}; 
    \input{sections/results/vae/cvae-mnist-distinct.tex}
	\end{axis}
\end{tikzpicture}}
\caption{MNIST, no overlap}
\end{subfigure}%
\hspace{10pt}
\begin{subfigure}[t]{.24\textwidth}
\resizebox{\textwidth}{!}{\begin{tikzpicture}
	\begin{axis}[
    xmode=log, legend pos=south east,
    width=2\textwidth,
    log basis x={2},
    legend style={font=\fontsize{3}{5}\selectfont},
	scatter/classes={%
		a={mark=square*,blue},%
		b={mark=triangle*,red},%
		c={mark=o,draw=black}}]
	
	\addplot[black,dashed,mark=o,mark options={solid}] coordinates {
	    (1, 0.19217608695652175)
	    (2, 0.2519413043478261)
	    (4, 0.32359130434782607)
	    (8, 0.38644347826086967)
	    (16, 0.4453260869565217)
	    (32, 0.5013282608695652)
	    (64, 0.5616326086956522)
	};
	\addlegendentry{SVM}
	\addlegendentry{+dVAE}
	\legend{}; 
    \input{sections/results/vae/cvae-emnist-overlap.tex}
	\end{axis}
\end{tikzpicture}}
\caption{EMNIST, overlap}
\end{subfigure}%
\begin{subfigure}[t]{.24\textwidth}
\resizebox{\textwidth}{!}{\begin{tikzpicture}
	\begin{axis}[
    legend cell align={left},
    xmode=log,
    legend pos=south east,
    width=2\textwidth,
	scatter/classes={%
		a={mark=square*,blue},%
		b={mark=triangle*,red},%
		c={mark=o,draw=black}}]
	
	\addplot[black,dashed,mark=o,mark options={solid}] coordinates {
        (1, 0.30927333333333334)
	    (2, 0.42175333333333337)
	    (4, 0.5314266666666667)
	    (8, 0.6244999999999999)
	    (16, 0.6951333333333334)
	    (32, 0.7513000000000001)
	    (64, 0.7939800000000001)
	};
	\addlegendentry{SVM}
	
	\addlegendentry{dVAE+SVM}
    \input{sections/results/vae/cvae-emnist-distinct.tex}
	\end{axis}
\end{tikzpicture}}
\caption{EMNIST, no overlap}
\end{subfigure}

\caption{Test accuracy obtained on MNIST and EMNIST. Average accuracy and
standard deviation are reported; to help readability, we did not include
the standard deviation of the SVMs.}
\label{fig:vae-cnn}
\end{figure}

Next, we compare accuracy of the Gaussian kernel SVM trained on the
original images versus the representation retrieved from dVAE.
Experiments are repeated $25$ times and averages are shown in
fig.~\ref{fig:vae-cnn}.
We can see that the classifier
indeed benefits from the representation of the dVAE in comparison to the
original images. However, when the class distributions do not overlap, the
classifier trained on the embeddings performs only marginally better on MNIST. As EMNIST has 37 more classes than MNIST, transfer learning appears to be more successful as the representation may be less specific and useful for the hold out classes. These preliminary results support the viability of our private protocol for transfer learning, although experiments on more realistic datasets and task shall be conducted in future work.

\section{Conclusions}

We have introduced SEALion, a new exensible framework for building deep neural networks that can perform inference on homomorphically encrypted data, protecting the privacy of data owners utilizing prediction services.
On top of the innovation of CryptoNets \cite{gilad2016cryptonets}, our contribution provides an modular and extensible software architecture for quick prototyping machine learning research ideas.
The automatic encryption parameters search algorithm side-steps many of the tedious implementation details that required ad-hoc solutions in previous work.

Built on SEALion, we also introduced two practically-relevant novelties in the area. We showed how to improve both latency and throughput of encrypted inference by sparsifying the neural networks' activations. Furthermore, we have defined the new scenario of encrypted transfer learning, where a server  exposes a feature extractor that preserves the privacy of the client data.

We believe that our contributions will benefit research efforts at the intersection of machine learning and homomorphic encryption, by providing both the tools and inspiration that are necessary to incorporate privacy-preserving aspects into the inference phase of machine learning models.

\bibliography{example_paper}

\begin{thebibliography}{31}
\providecommand{\natexlab}[1]{#1}
\providecommand{\url}[1]{\texttt{#1}}
\expandafter\ifx\csname urlstyle\endcsname\relax
  \providecommand{\doi}[1]{doi: #1}\else
  \providecommand{\doi}{doi: \begingroup \urlstyle{rm}\Url}\fi

\bibitem[Bengio(2012)]{bengio2012deep}
Bengio, Y.
\newblock Deep learning of representations for unsupervised and transfer
  learning.
\newblock In \emph{Proceedings of ICML Workshop on Unsupervised and Transfer
  Learning}, pp.\  17--36, 2012.

\bibitem[Bengio et~al.(2013)Bengio, L{\'e}onard, and
  Courville]{bengio2013estimating}
Bengio, Y., L{\'e}onard, N., and Courville, A.
\newblock Estimating or propagating gradients through stochastic neurons for
  conditional computation.
\newblock 2013.

\bibitem[Bos et~al.(2013)Bos, Lauter, Loftus, and Naehrig]{bos2013improved}
Bos, J.~W., Lauter, K., Loftus, J., and Naehrig, M.
\newblock Improved security for a ring-based fully homomorphic encryption
  scheme.
\newblock In \emph{IMA International Conference on Cryptography and Coding},
  pp.\  45--64. Springer, 2013.

\bibitem[Bourse et~al.(2018)Bourse, Minelli, Minihold, and
  Paillier]{bourse2018fast}
Bourse, F., Minelli, M., Minihold, M., and Paillier, P.
\newblock Fast homomorphic evaluation of deep discretized neural networks.
\newblock In \emph{Annual International Cryptology Conference}, pp.\  483--512.
  Springer, 2018.

\bibitem[Chase et~al.(2017)Chase, Chen, Ding, Goldwasser, Gorbunov, Hoffstein,
  Lauter, Lokam, Moody, Morrison, et~al.]{chase2017security}
Chase, M., Chen, H., Ding, J., Goldwasser, S., Gorbunov, S., Hoffstein, J.,
  Lauter, K., Lokam, S., Moody, D., Morrison, T., et~al.
\newblock Security of homomorphic encryption.
\newblock \emph{HomomorphicEncryption. org, Redmond WA, Tech. Rep}, 2017.

\bibitem[Chillotti et~al.(2016)Chillotti, Gama, Georgieva, and
  Izabachene]{chillotti2016faster}
Chillotti, I., Gama, N., Georgieva, M., and Izabachene, M.
\newblock Faster fully homomorphic encryption: Bootstrapping in less than 0.1
  seconds.
\newblock In \emph{International Conference on the Theory and Application of
  Cryptology and Information Security}, pp.\  3--33. Springer, 2016.

\bibitem[Chou et~al.(2018)Chou, Beal, Levy, Yeung, Haque, and
  Fei-Fei]{chou2018faster}
Chou, E., Beal, J., Levy, D., Yeung, S., Haque, A., and Fei-Fei, L.
\newblock Faster cryptonets: Leveraging sparsity for real-world encrypted
  inference.
\newblock \emph{arXiv preprint arXiv:1811.09953}, 2018.

\bibitem[Cohen et~al.(2017)Cohen, Afshar, Tapson, and van
  Schaik]{cohen2017emnist}
Cohen, G., Afshar, S., Tapson, J., and van Schaik, A.
\newblock Emnist: an extension of mnist to handwritten letters.
\newblock \emph{arXiv preprint arXiv:1702.05373}, 2017.

\bibitem[Courbariaux et~al.(2016)Courbariaux, Hubara, Soudry, El-Yaniv, and
  Bengio]{courbariaux2016binarized}
Courbariaux, M., Hubara, I., Soudry, D., El-Yaniv, R., and Bengio, Y.
\newblock {Binarized neural networks: Training deep neural networks with
  weights and activations constrained to+ 1 or-1}.
\newblock 2016.

\bibitem[Dahl et~al.(2018)Dahl, Mancuso, Dupis, Decoste, Giraud, Livingstone,
  Patriquin, and Uhma]{dahl2018private}
Dahl, M., Mancuso, J., Dupis, Y., Decoste, B., Giraud, M., Livingstone, I.,
  Patriquin, J., and Uhma, G.
\newblock Private machine learning in tensorflow using secure computation.
\newblock \emph{arXiv preprint arXiv:1810.08130}, 2018.

\bibitem[Dathathri et~al.(2018)Dathathri, Saarikivi, Chen, Laine, Lauter,
  Maleki, Musuvathi, and Mytkowicz]{dathathri2018chet}
Dathathri, R., Saarikivi, O., Chen, H., Laine, K., Lauter, K., Maleki, S.,
  Musuvathi, M., and Mytkowicz, T.
\newblock Chet: Compiler and runtime for homomorphic evaluation of tensor
  programs.
\newblock \emph{arXiv preprint arXiv:1810.00845}, 2018.

\bibitem[Dowlin et~al.(2017)Dowlin, Gilad-Bachrach, Laine, Lauter, Naehrig, and
  Wernsing]{dowlin2017manual}
Dowlin, N., Gilad-Bachrach, R., Laine, K., Lauter, K., Naehrig, M., and
  Wernsing, J.
\newblock Manual for using homomorphic encryption for bioinformatics.
\newblock \emph{Proceedings of the IEEE}, 105\penalty0 (3):\penalty0 552--567,
  2017.

\bibitem[Fan \& Vercauteren(2012)Fan and Vercauteren]{fan2012somewhat}
Fan, J. and Vercauteren, F.
\newblock Somewhat practical fully homomorphic encryption.
\newblock \emph{IACR Cryptology ePrint Archive}, 2012:\penalty0 144, 2012.

\bibitem[Gilad-Bachrach et~al.(2016)Gilad-Bachrach, Dowlin, Laine, Lauter,
  Naehrig, and Wernsing]{gilad2016cryptonets}
Gilad-Bachrach, R., Dowlin, N., Laine, K., Lauter, K., Naehrig, M., and
  Wernsing, J.
\newblock Cryptonets: Applying neural networks to encrypted data with high
  throughput and accuracy.
\newblock In \emph{International Conference on Machine Learning}, pp.\
  201--210, 2016.

\bibitem[Hardy et~al.(2017)Hardy, Henecka, Ivey-Law, Nock, Patrini, Smith, and
  Thorne]{hardy2017private}
Hardy, S., Henecka, W., Ivey-Law, H., Nock, R., Patrini, G., Smith, G., and
  Thorne, B.
\newblock Private federated learning on vertically partitioned data via entity
  resolution and additively homomorphic encryption.
\newblock 2017.

\bibitem[Juvekar et~al.(2018)Juvekar, Vaikuntanathan, and
  Chandrakasan]{juvekar2018gazelle}
Juvekar, C., Vaikuntanathan, V., and Chandrakasan, A.
\newblock Gazelle: A low latency framework for secure neural network inference.
\newblock 2018.

\bibitem[Kingma \& Ba(2014)Kingma and Ba]{kingma2014adam}
Kingma, D.~P. and Ba, J.
\newblock Adam: A method for stochastic optimization.
\newblock \emph{arXiv preprint arXiv:1412.6980}, 2014.

\bibitem[Kingma \& Welling(2013)Kingma and Welling]{kingma2013auto}
Kingma, D.~P. and Welling, M.
\newblock Auto-encoding variational bayes.
\newblock 2013.

\bibitem[Kingma et~al.(2014)Kingma, Mohamed, Rezende, and
  Welling]{kingma2014semi}
Kingma, D.~P., Mohamed, S., Rezende, D.~J., and Welling, M.
\newblock Semi-supervised learning with deep generative models.
\newblock In \emph{Advances in Neural Information Processing Systems}, pp.\
  3581--3589, 2014.

\bibitem[Kononenko(2001)]{kononenko2001machine}
Kononenko, I.
\newblock Machine learning for medical diagnosis: history, state of the art and
  perspective.
\newblock \emph{Artificial Intelligence in medicine}, 23\penalty0 (1):\penalty0
  89--109, 2001.

\bibitem[Laine \& Player(2016)Laine and Player]{laine2016seal}
Laine, K. and Player, R.
\newblock Simple encrypted arithmetic library-seal.
\newblock Technical report, Technical report, September, 2016.

\bibitem[LeCun et~al.(1998)LeCun, Bottou, Bengio, and
  Haffner]{lecun1998gradient}
LeCun, Y., Bottou, L., Bengio, Y., and Haffner, P.
\newblock Gradient-based learning applied to document recognition.
\newblock \emph{Proceedings of the IEEE}, 86\penalty0 (11):\penalty0
  2278--2324, 1998.

\bibitem[Louizos et~al.(2017)Louizos, Welling, and Kingma]{louizos2017learning}
Louizos, C., Welling, M., and Kingma, D.~P.
\newblock {Learning Sparse Neural Networks through $L_0$ Regularization}.
\newblock 2017.

\bibitem[Maddison et~al.(2016)Maddison, Mnih, and Teh]{maddison2016concrete}
Maddison, C.~J., Mnih, A., and Teh, Y.~W.
\newblock The concrete distribution: A continuous relaxation of discrete random
  variables.
\newblock \emph{arXiv preprint arXiv:1611.00712}, 2016.

\bibitem[Makri et~al.(2019)Makri, Rotaru, Smart, and Vercauteren]{makriepic}
Makri, E., Rotaru, D., Smart, N.~P., and Vercauteren, F.
\newblock Epic: Efficient private image classification (or: Learning from the
  masters).
\newblock \emph{To appear in CT-RSA}, 2019.

\bibitem[Noroozi \& Favaro(2016)Noroozi and Favaro]{noroozi2016unsupervised}
Noroozi, M. and Favaro, P.
\newblock Unsupervised learning of visual representations by solving jigsaw
  puzzles.
\newblock In \emph{European Conference on Computer Vision}, pp.\  69--84.
  Springer, 2016.

\bibitem[Pathak et~al.(2016)Pathak, Krahenbuhl, Donahue, Darrell, and
  Efros]{pathak2016context}
Pathak, D., Krahenbuhl, P., Donahue, J., Darrell, T., and Efros, A.~A.
\newblock Context encoders: Feature learning by inpainting.
\newblock In \emph{Proceedings of the IEEE Conference on Computer Vision and
  Pattern Recognition}, pp.\  2536--2544, 2016.

\bibitem[Rezende et~al.(2014)Rezende, Mohamed, and
  Wierstra]{rezende2014stochastic}
Rezende, D.~J., Mohamed, S., and Wierstra, D.
\newblock Stochastic backpropagation and approximate inference in deep
  generative models.
\newblock 2014.

\bibitem[Ryffel et~al.(2018)Ryffel, Trask, Dahl, Wagner, Mancuso, Rueckert, and
  Passerat-Palmbach]{ryffel2018generic}
Ryffel, T., Trask, A., Dahl, M., Wagner, B., Mancuso, J., Rueckert, D., and
  Passerat-Palmbach, J.
\newblock A generic framework for privacy preserving deep learning.
\newblock \emph{arXiv preprint arXiv:1811.04017}, 2018.

\bibitem[Sanyal et~al.(2018)Sanyal, Kusner, Gasc{\'o}n, and
  Kanade]{sanyal2018tapas}
Sanyal, A., Kusner, M.~J., Gasc{\'o}n, A., and Kanade, V.
\newblock {TAPAS: Tricks to Accelerate (encrypted) Prediction As a Service}.
\newblock 2018.

\bibitem[Smart \& Vercauteren(2014)Smart and Vercauteren]{smart2014fully}
Smart, N.~P. and Vercauteren, F.
\newblock Fully homomorphic simd operations.
\newblock \emph{Designs, codes and cryptography}, 71\penalty0 (1):\penalty0
  57--81, 2014.

\end{thebibliography}
\bibliographystyle{icml2019}

\end{document}